\documentclass[10pt]{article} % For LaTeX2e
\usepackage[preprint]{tmlr}
\usepackage{amsmath,amsfonts,bm}

\def\eqref#1{equation~\ref{#1}}
\def\1{\bm{1}}

\DeclareMathAlphabet{\mathsfit}{\encodingdefault}{\sfdefault}{m}{sl}
\SetMathAlphabet{\mathsfit}{bold}{\encodingdefault}{\sfdefault}{bx}{n}

\usepackage{hyperref}
\usepackage{url}
\usepackage{amssymb}
\usepackage{amsmath}
\usepackage{amsthm}
\usepackage{comment}
\usepackage{booktabs}       % professional-quality tables
\usepackage{amsfonts}       % blackboard math symbols

\usepackage{mathtools}
\usepackage{algorithm}
\usepackage{multirow}

\usepackage{hyperref}
\usepackage{url}
\usepackage{amsthm}
\usepackage{placeins}

\newcommand{\MMD}{\operatorname{MMD}}
\newcommand{\CORAL}{\operatorname{CORAL}}
\usepackage{algorithm}
\usepackage{algpseudocode}
\usepackage{graphicx}
\usepackage{caption}
\usepackage{subcaption}
\usepackage{booktabs}       % professional-quality tables

\title{Online Variance Reduction for Domain Adaptation on Streaming Data}

\author{\name Andrea Napoli  \\
      \addr Department of Electronics and Computer Science\\
  University of Southampton, UK
}

\def\month{05}  % Insert correct month for camera-ready version
\def\year{2026} % Insert correct year for camera-ready version

\begin{document}

\maketitle

\begin{abstract}
This paper studies the problem of stochastic variance reduction (SVR) for the maximum mean discrepancy (MMD) and correlation alignment (CORAL) loss functions. Although various offline SVR algorithms for these losses have been proposed, these are incompatible with online, distributed, or incremental learning settings. This paper presents Adaptive vaRiance Reduction via Online reWeighting (ARROW), the first online SVR algorithm for the MMD and CORAL for streamed data. The method maintains moving average references of the alignment statistics, and adaptively reweights incoming minibatches so that the minibatch and reference statistics are aligned. Further, we propose a relaxed reweighting scheme so that the ensuing weight-optimisation problem is tractable. In experiments and simulations, we show that ARROW performs competitively with offline algorithms in terms of runtime, degree of variance reduction achieved, and target domain accuracy.
\end{abstract}

\section{Introduction}

Domain shift remains a key challenge when deploying machine learning models to the real world \citep{Gulrajani2021InGeneralization,Koh2021WILDS:Shifts}. Unsupervised domain adaptation (UDA) aims to address this by minimising a “domain discrepancy” loss during training, which characterises the mismatch between the source and target feature distributions. Two widely used objectives are the correlation alignment (CORAL) loss, which matches covariance matrices \citep{Sun2016DeepAdaptation}, and the maximum mean discrepancy (MMD), which aligns kernel mean embeddings \citep{Tzeng2014DeepInvariance,pmlr-v37-long15,Li2018DomainLearning}.

Although theoretically well-grounded \citep{Ben-David2006AnalysisAdaptation,Ben-David2010ADomains,Redko2022AGuarantees}, estimating these discrepancies in practice is extremely noisy, especially in minibatch optimisation settings. This high variance can destabilise training, and these methods have frequently been observed to perform worse than with no domain alignment at all \citep{Dubey2021AdaptiveGeneralization,Gao2023Out-of-DistributionAugmentations,Gulrajani2021InGeneralization,Koh2021WILDS:Shifts,Napoli2023UnsupervisedCalls,Napoli2024ImprovingSampling,Wang2019CharacterizingTransfer}. Consequentially, a recent paper found that stochastic variance reduction (SVR) can function as an effective stabiliser for these methods, resulting in significant gains in target domain performance \citep{Napoli2026VarianceSampling}. However, most classical SVR methods require the losses to possess finite-sum structure, which the MMD and CORAL terms lack.

Thus, a line of research has emerged developing SVR techniques which are compatible with, or even specialised for, the kinds of non-additive losses found in UDA. The first work to do so proposed diverse sampling of minibatches \citep{Napoli2024ImprovingSampling}; this also helps to balance the data \citep{Napoli2024Diversity-BasedAdaptation}. Outside of UDA, diverse sampling has also previously been found to be beneficial to other nonadditive objectives, such as contrastive losses \citep{Wu2023ContrastiveSamples,Ochieng2025DiversityMagnitudes}, as well as minibatch optimisation more broadly \citep{Zhang2017DeterminantalDiversification,Zhang2019ActiveProcesses,Bardenet2021DeterminantalSGD,Jiang2024DOS:Detection}. However, as well as being computationally rather slow, the effectiveness of this approach is limited by the relatively weak connection between diversity and variance.

The first dedicated SVR technique for UDA was VaRDASS \citep{Napoli2026VarianceSampling}, which leveraged stratified sampling. VaRDASS forms the strata using discrepancy-specific clustering objectives, which directly minimises variance for the MMD and CORAL (with respect to the strata). The computational cost is also negligible once the strata are constructed. Stratified sampling has also been previously explored for minibatch optimisation outside of UDA \citep{Zhao2014AcceleratingSampling,Liu2020AcceleratingStrata,Fu2017CPSG-MCMC:MCMC,Lu2021VarianceModels}.

Diverse and stratified sampling both reduce variance by inducing negative correlation between observations within domains. However, the variance can be reduced further by inducing positive correlation across domains, as well as dependencies across training steps.
To achieve this, \citet{Napoli2026OrderData} proposed ORDERED, which directly optimises variance with respect to the training data sampling order. ORDERED starts with a random data sequence and then iteratively exchanges datapoints between minibatches such that the variance is reduced. This is a type of deterministic minibatch sequencing, which also includes techniques such as curriculum learning \citep{Bengio2009CurriculumLearning} and anti-clustering \citep{Papenberg2021UsingParts,Baumann2026AAlgorithm}. However, the improvement in variance reduction comes at significant computational cost. Additionally, the optimisation procedure relies on a greedy heuristic which is sensitive to local minima, so the solution quality also retains room for improvement.

Most recently, \citet{Napoli2026Variance-reducedSampling} proposed PSDA, which constructs sampling dependencies between datapoints so as to reduce the expected discrepancy gradient error, conditional on those datapoints being jointly observed. The dependencies are constructed in pairs using linear assignments, rendering the method fast and scalable. The underlying idea, known as paired or antithetic sampling, has also been explored for minibatch optimisation outside of UDA \citep{Liu2018AcceleratingSampling}.

All these preceding methods rely on modifying the data sampling process. Although this has advantages in terms of convenience, there are also certain disadvantages. Firstly, the datasets need to be fully accessible at all times before and during training. This is unacceptable in many settings, such as online, distributed, or incremental learning paradigms -- and is especially problematic in UDA, where continual adaptation to domain shift is often required. Secondly, features for the full dataset are periodically re-extracted throughout training, which can entail significant overhead cost. Lastly, even linearly-scaling implementations of these algorithms may still be too slow for extremely large datasets.

This paper presents Adaptive vaRiance Reduction via Online reWeighting (ARROW), the first online SVR algorithm specialised for the MMD and CORAL. The method maintains exponentially weighted moving averages (EWMAs) of the adaptation statistics (kernel mean embeddings and covariances respectively), and aligns the minibatch statistics with the EWMA references using weighted losses. The resulting minibatch gradients are thus aligned to a variance-reduced “full” gradient implied by the reference statistics. As well as not requiring the full dataset, the use of EWMAs allows the reference statistics to track the true population statistics as the latter vary during training, or as the input distributions shift. Further, we propose relaxed reweighting schemes based on linear combinations, yielding unconstrained quadratic weight-optimisation problems that can be readily solved using standard techniques.

In experiments and simulations, we assess ARROW in terms of runtime, degree of variance reduction achieved, and target domain accuracy, demonstrating competitiveness with offline algorithms on all three criteria.

\section{Method}\label{method}

\subsection{Preliminaries}\label{preliminaries}

Let \(x_{s},y_{s}\) denote the source input data and labels and
\(x_{t}\) the unlabelled target data. We assume a model \(h\) comprising
a feature extractor \(f\) and prediction head \(g\), such that
\(h = g \circ f\), with corresponding feature representations
\(z_{s} = f\left( x_{s} \right),\ \ z_{t} = f\left( x_{t} \right)\).
Since the target data are unlabelled, UDA optimises a supervised task
loss on the source data, together with a distribution-matching
regulariser which aligns the two feature distributions:
\begin{equation}L = L_{\mathrm{task}}\left( h\left( x_{s} \right),y_{s} \right) + \lambda L_{\mathrm{disc}}\left( z_{s},z_{t} \right),\end{equation}
where \(\lambda > 0\) is a trade-off parameter.

This paper considers two specific options for \(L_{\mathrm{disc}}\), the MMD and
CORAL. The (squared) MMD loss is defined as
\begin{equation}L_{\MMD} = \left\| \mu_{s} - \mu_{t} \right\|_{\mathcal{H}}^{2},\end{equation}
where
\(\mu_{s}\mathbb{= E}\left\lbrack \phi\left( z_{s} \right) \right\rbrack,\mu_{t}\mathbb{= E}\left\lbrack \phi\left( z_{t} \right) \right\rbrack\),
\(\mathcal{H}\) is a reproducing kernel Hilbert space (RKHS), and
\(\phi\ :\mathcal{Z \rightarrow H}\) is an implicit mapping.
\(\mathcal{H}\) is associated with a unique positive-definite kernel
\(\kappa:\mathcal{Z \times Z}\mathbb{\rightarrow R}\) for which the
reproducing property
\(\kappa(z,z') = \left\langle \phi(z),\phi(z') \right\rangle_{\mathcal{H}}\)
is satisfied. On the other hand, CORAL aims to minimise the (squared)
Frobenius distance between the source and target feature covariance
matrices:
\begin{equation}L_{\CORAL} = \left\| \Sigma_{s} - \Sigma_{t} \right\|_{F}^{2}.\end{equation}
Define the kernel mean difference \(D_{\MMD} = \mu_{s} - \mu_{t}\) and
covariance difference \(D_{\CORAL} = \Sigma_{s} - \Sigma_{t}\). We will
also use \(D\) to refer generically to both quantities, and
\(\left\| \cdot \right\|\) for the associated norm.

We consider a data streaming setting in which the training data arrive
in minibatches over time (from potentially time-varying distributions).
We will assume for simplicity that all minibatches share the same size
\(k\). Specifically, at timestep \(m\), we receive training data
\(\left\{ \left( x_{s,i}^{m},y_{s,i}^{m} \right) \right\}_{i = 1}^{k}\)
and \(\left\{ x_{t,j}^{m} \right\}_{j = 1}^{k}\), and compute stochastic
losses \({\widehat{L}}_{\mathrm{task}}^{m}\) and \({\widehat{L}}_{\mathrm{disc}}^{m}\).
For both the MMD and CORAL, we use the V-statistic estimators
\({\widehat{L}}_{\mathrm{disc}}^{m} = \left\| {\widehat{D}}^{m} \right\|^{2}\),
with
\({\widehat{D}}_{\MMD}^{m} = {\widehat{\mu}}_{s}^{m} - {\widehat{\mu}}_{t}^{m}\)
and
\({\widehat{D}}_{\CORAL}^{m} = {\widehat{\Sigma}}_{s}^{m} - {\widehat{\Sigma}}_{t}^{m}\)
respectively.

We then compute stochastic gradients \(\widehat{\nabla}L_{\mathrm{disc}}^{m}\)
and \(\widehat{\nabla}L_{\mathrm{task}}^{m}\), which are noisy estimates of the
corresponding full gradients. Since the discrepancy gradients are the
main source of instability during training, our aim is to use SVR to
reduce the discrepancy gradient error at each timestep,
\(\left\| \widehat{\nabla}L_{\mathrm{disc}}^{m} - \nabla L_{\mathrm{disc}}^{m} \right\|^{2}\).
Since, by the chain rule,
\(\widehat{\nabla}L_{\mathrm{disc}}^{m} - \nabla L_{\mathrm{disc}}^{m} \propto {\widehat{D}}^{m} - D^{m}\),
we can do this by minimising
\(\left\| {\widehat{D}}^{m} - D^{m} \right\|^{2}\) directly (subject to some smoothness conditions discussed in \citet{Zhao2014AcceleratingSampling,Liu2020AcceleratingStrata}). For ARROW,
this entails substituting an instance-weighted difference
\({\widehat{D}}_{w}^{m}\), where the weights are optimised at each
training step. In addition, since \(D^{m}\) is unknown and time-varying,
we instead align to an EWMA estimate \({\widetilde{D}}^{m}\) which is
maintained online:
\begin{equation}\min\left\| {\widehat{D}}_{w}^{m} - {\widetilde{D}}^{m} \right\|^{2}.\end{equation}
The procedure differs for the MMD and CORAL.

\subsection{MMD}\label{mmd}

The EWMA of the kernel mean difference can be represented by the
recurrence relation
\begin{gather}{\widetilde{D}}_{\MMD}^{0} = {\widehat{D}}_{\MMD}^{0} \\{\widetilde{D}}_{\MMD}^{m} = (1 - \alpha){\widetilde{D}}_{\MMD}^{m - 1} + \alpha{\widehat{D}}_{\MMD}^{m},\end{gather}
where \(\alpha \in (0,1\rbrack\) is the decay factor. However, since the
explicit RKHS embeddings are unavailable, this needs to be computed by
expanding the recurrence into a closed-form summation and storing a
buffer of historical minibatches, for which the cross-kernels with the
current minibatch can be evaluated. Minibatches with sufficiently small
coefficients can be discarded to control memory growth.

We define \({\widehat{D}}_{\MMD,w}^{m}\) as a linear combination of RKHS
embeddings:
\begin{gather}{\widehat{D}}_{\MMD,w}^{m} = {\widehat{\mu}}_{s,w}^{m} - {\widehat{\mu}}_{t,w}^{m},\\
{\widehat{\mu}}_{s,w}^{m} = \sum_{i = 1}^{k}{u_{i}\phi\left( z_{s,i}^{m} \right)},\quad {\widehat{\mu}}_{t,w}^{m} = \sum_{j = 1}^{k}{v_{j}\phi\left( z_{t,j}^{m} \right)},\end{gather}
with weight vectors \(u,v \in \mathbb{R}^{k}\) (note that we are
omitting the \(m\) superscripts here for notational convenience). Unlike
in KMM, where the kernel means are convex combinations of RKHS
embeddings, we do not impose any constraints on \(u\) and \(v\), since
the weights do not need to resemble a probability distribution (thus
\({\widehat{\mu}}_{s,w}^{m}\) and \({\widehat{\mu}}_{t,w}^{m}\) are
technically no longer ``means''). This increases the power of our
optimisation: with arbitrary weights, we can perfectly align
\(\left\| {\widehat{D}}_{\MMD,w}^{m} - {\widetilde{D}}_{\MMD}^{m} \right\|_{\mathcal{H}}^{2} = 0\)
as long as \({\widetilde{D}}_{MMD}^{m}\) is contained in the linear span
(as opposed to the convex hull) of \(\phi\left( z_{s,i}^{m} \right)\)
and \(\phi\left( z_{t,j}^{m} \right)\).

We can obtain the required objective function in terms of kernel
evaluations by separating
\begin{equation}\left\| {\widehat{D}}_{\MMD,w}^{m} - {\widetilde{D}}_{\MMD}^{m} \right\|_{\mathcal{H}}^{2} = \left\| {\widehat{\mu}}_{s,w}^{m} - {\widehat{\mu}}_{t,w}^{m} \right\|_{\mathcal{H}}^{2} + \left\| {\widetilde{D}}_{\MMD}^{m} \right\|_{\mathcal{H}}^{2} - 2\left\langle {\widehat{\mu}}_{s,w}^{m} - {\widehat{\mu}}_{t,w}^{m},{\widetilde{D}}_{\MMD}^{m} \right\rangle_{\mathcal{H}}.\end{equation}
First, we observe that
\(\left\| {\widetilde{D}}_{\MMD}^{m} \right\|_{\mathcal{H}}^{2}\) is a
constant that does not affect the optimisation. Then, we have
\begin{gather}\left\| {\widehat{\mu}}_{s,w}^{m} - {\widehat{\mu}}_{t,w}^{m} \right\|_{\mathcal{H}}^{2} = u^{T}K_{ss}^{m}u + v^{T}K_{tt}^{m}v - 2u^{T}K_{st}^{m}v,\\
K_{ss,i,j}^{m} = \kappa\left( z_{s,i}^{m},z_{s,j}^{m} \right), \quad
K_{tt,i,j}^{m} = \kappa\left( z_{t,i}^{m},z_{t,j}^{m} \right), \quad
K_{st,i,j}^{m} = \kappa\left( z_{s,i}^{m},z_{t,j}^{m} \right),\end{gather}
and
\begin{gather}\left\langle {\widehat{\mu}}_{s,w}^{m} - {\widehat{\mu}}_{t,w}^{m},{\widetilde{D}}_{\MMD}^{m} \right\rangle_{\mathcal{H}} = u^{T}b_{s} - v^{T}b_{t},\\
b_{s,i} = \alpha\sum_{l = 1}^{m}{(1 - \alpha)^{m - l}\sum_{j = 1}^{k}{\kappa\left( z_{s,i}^{m},z_{s,j}^{l} \right) - \kappa\left( z_{s,i}^{m},z_{t,j}^{l} \right)}} + (1 - \alpha)^{m}\sum_{j = 1}^{k}{\kappa\left( z_{s,i}^{m},z_{s,j}^{0} \right) - \kappa\left( z_{s,i}^{m},z_{t,j}^{0} \right)},\\
b_{t,i} = \alpha\sum_{l = 1}^{m}{(1 - \alpha)^{m - l}\sum_{j = 1}^{k}{\kappa\left( z_{t,i}^{m},z_{s,j}^{l} \right) - \kappa\left( z_{t,i}^{m},z_{t,j}^{l} \right)}} + (1 - \alpha)^{m}\sum_{j = 1}^{k}{\kappa\left( z_{t,i}^{m},z_{s,j}^{0} \right) - \kappa\left( z_{t,i}^{m},z_{t,j}^{0} \right)}.\end{gather}
Since the kernel is positive-definite by definition, we thus have the
convex quadratic optimisation
\begin{equation}\min_{u,v}{u^{T}K_{ss}^{m}u + v^{T}K_{tt}^{m}v - 2u^{T}K_{st}^{m}v - 2\left( u^{T}b_{s} - v^{T}b_{t} \right)},\end{equation}
which can be solved efficiently using standard methods.

\subsection{CORAL}\label{coral}

For CORAL, we need to maintain EWMAs of both the means and covariances
to correctly update \({\widetilde{D}}_{\CORAL}^{m}\). The mean updates
are
\begin{gather}{\widetilde{c}}_{s}^{0} = {\widehat{c}}_{s}^{0},\\
{\widetilde{c}}_{s}^{m} = (1 - \alpha){\widetilde{c}}_{s}^{m - 1} + \alpha{\widehat{c}}_{s}^{m},\end{gather}
and the covariances are given by
\begin{gather}{\widetilde{\Sigma}}_{s}^{0} = {\widehat{\Sigma}}_{s}^{0},\\{\widetilde{\Sigma}}_{s}^{m} = (1 - \alpha){\widetilde{\Sigma}}_{s}^{m - 1} + \alpha{\widehat{\Sigma}}_{s}^{m} + (1 - \alpha)\alpha\left( {\widehat{c}}_{s}^{m} - {\widetilde{c}}_{s}^{m - 1} \right)\left( {\widehat{c}}_{s}^{m} - {\widetilde{c}}_{s}^{m - 1} \right)^{T},\end{gather}
where the latter term corrects for the shift in mean between updates.

We now need weighted ``covariance'' expressions which can be plugged
into \({\widehat{D}}_{\CORAL,w}^{m}\). However, using the standard
weighted covariance formulae would yield a quartic weight-optimisation problem, which is NP-hard. Therefore, we propose instead a
relaxation comprising linear combinations of outer products:
\begin{gather}{\widehat{D}}_{\CORAL,w}^{m} = {\widehat{\Sigma}}_{s,w}^{m} - {\widehat{\Sigma}}_{t,w}^{m},\\
{\widehat{\Sigma}}_{s,w}^{m} = \sum_{i = 1}^{k}{u_{i}\left( z_{s,i}^{m} - {\widehat{c}}_{s}^{m} \right)\left( z_{s,i}^{m} - {\widehat{c}}_{s}^{m} \right)^{T}},\\
{\widehat{\Sigma}}_{t,w}^{m} = \sum_{j = 1}^{k}{v_{j}\left( z_{t,j}^{m} - {\widehat{c}}_{t}^{m} \right)\left( z_{t,j}^{m} - {\widehat{c}}_{t}^{m} \right)^{T}},\end{gather}
with fixed (unweighted) centres
\({\widehat{c}}_{s}^{m},{\widehat{c}}_{t}^{m}\). Thus, we obtain a
tractable linear least squares problem
\begin{equation}\min_{u,v}\left\| {\widehat{D}}_{\CORAL,w}^{m} - {\widetilde{D}}_{\CORAL}^{m} \right\|_{F}^{2},\end{equation}
which is solvable to 0 if the linear span of the outer products contains
\({\widetilde{D}}_{\CORAL}^{m}\).

\section{Experiments}\label{experiments}

In this section, we evaluate ARROW on three criteria: degree of variance
reduction achieved, target domain accuracy, and training speed. For all experiments, we set $\alpha=0.1$, and limit the MMD minibatch buffer to coefficients above 0.01 (so storing a maximum of 22 minibatches).

\subsection{Estimator variance}\label{estimator-variance}

\begin{figure}
    \centering
    \includegraphics[width=0.5\linewidth]{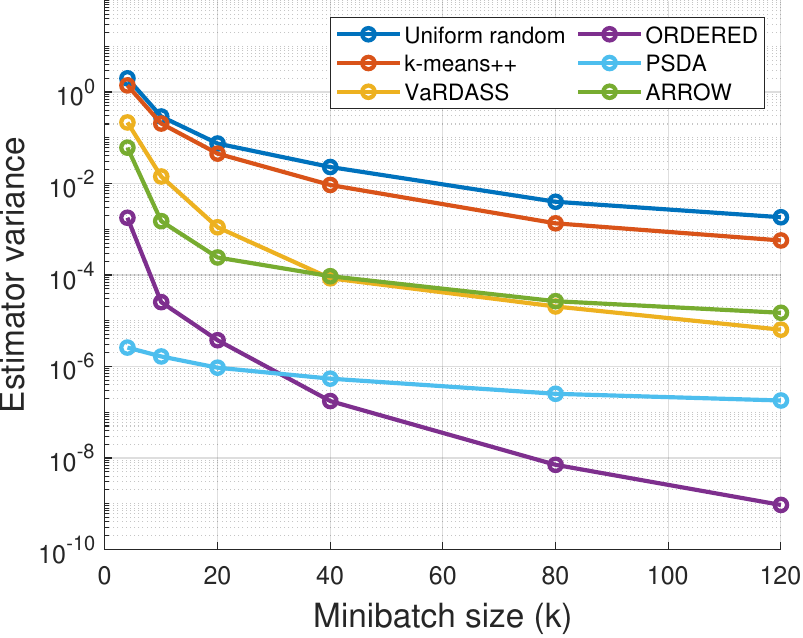}
    \caption{Estimator variance vs $k$ for six sampling algorithms.}
    \label{var}
\end{figure}

We use Monte Carlo simulations to compare the estimator variance for
different SVR methods across different values of \(k\) (Figure \ref{var}). Specifically, we
compute \({\mathbb{E}\left\| \widehat{D} - D \right\|}^{2}\) using a
linear kernel (that is, estimating the squared Euclidean distance
between distribution means) between a source and target dataset
comprising 2D standard normal data with \(n_{s} = n_{t} = 4,000\). The experiment is run using 1,000 samples, and the EWMAs are updated with every sample as in realistic online conditions. In
addition to ARROW, we compare uniform random sampling,
diverse sampling using k-means++ \citep{Arthur2007K-means++:Seeding,Napoli2024ImprovingSampling}, stratified sampling (VaRDASS) \citep{Napoli2026VarianceSampling},
order-aware sampling (ORDERED) \citep{Napoli2026OrderData}, and paired sampling (PSDA) \citep{Napoli2026Variance-reducedSampling}. The graph shows that ARROW performs comparably to VaRDASS, slightly outperforming it for lower values of $k$.

\subsection{Target domain accuracy}\label{target-domain-accuracy}

Next, we evaluate ARROW in realistic training conditions, to assess
whether the observed reduction in variance translates to an increase in
test accuracy. Experiments are conducted
using the DomainBed framework \citep{Gulrajani2021InGeneralization} on the following domain shift
benchmarks.

\textbf{Spawrious} \citep{Lynch2023Spawrious:Biases} classification of 4 dog breeds
across images with different background environments (desert, jungle,
snow etc.). This benchmark comprises 6 data splits of varying difficulty, 6 domains and 18,664 examples.

\textbf{Office-Home} \citep{Venkateswara2017DeepAdaptation} image classification of 65 categories of everyday objects with different image styles (Art, Clipart, Product, and Real World). This benchmark comprises 12 data splits, 4 domains and 15,500 examples.

\textbf{Humpbacks} \citep{Napoli2023UnsupervisedCalls} detection of humpback whale
vocalisations in underwater acoustic recordings across data from
different acoustic monitoring programs. This benchmark comprises 4 data splits, 4 domains and 8,000 examples.

The domain discrepancies are measured between the union of all training
data and a held-out subset of the evaluation set. For the MMD, we use a
radial basis function (RBF) mixture kernel \citep{Li2018DomainLearning}, given by
\(\kappa(z,z') = \sum_{\gamma \in \mathcal{G}}^{}e^{- \gamma\left\| z - z' \right\|^{2}}\)with
\(\mathcal{G} = \{ 0.001,0.01,0.1,1,10\}\).
For Spawrious and Office-Home, we use a ResNet-18 \citep{He2015DeepRecognition} backbone pre-trained on ImageNet. For
Humpbacks, we use the audio frontend and architecture described in
\citet{Napoli2023UnsupervisedCalls}. Models are trained using the Adam optimiser \citep{Kingma2014Adam:Optimization} for 3,000
iterations. Hyperparameters, including the learning rate, weight decay, minibatch size $k$, and UDA trade-off parameter, are tuned with a random search of size 10
using an in-distribution (training domain) validation set, independently
for each sampler. In particular, the random search distribution for $k$ is $k\sim2^{\text{Uniform}(3,7)}$. The entire set of experiments is repeated 5 times for
reproducibility, using different random seeds for hyperparameters,
weight initialisations, and dataset splits. All other hyperparameter
choices and training details follow the DomainBed default options.

In addition to ARROW, six variance-reduced samplers are compared: k-means++, DPP \citep{Zhang2017DeterminantalDiversification,Napoli2024ImprovingSampling}, anticlustering \citep{Baumann2026AAlgorithm}, VaRDASS, ORDERED, and PSDA, as well as PSDA and ARROW together. We also compare several baseline UDA methods: DANN \citep{Ganin2015Domain-AdversarialNetworks}, CDAN \citep{Long2017ConditionalAdaptation}, SDAT \citep{Rangwani2022ATraining}, ELS \citep{Zhang2023FreeSmoothing}, ARM \citep{Zhang2020AdaptiveShift}, and MCC \citep{Jin2020MinimumAdaptation}, plus non-adaptive training via ERM \citep{Vapnik1998StatisticalTheory}.

Tables
\ref{spawrious split}, \ref{office split}, and \ref{humps split} show the average test accuracy and
standard errors over the 5 repeats and each of the data splits, for
each method. The results support the notion that SVR significantly boosts the performance of the MMD and CORAL for UDA. This allows these classical and simple techniques to outperform substantially more modern and complicated methods. Overall, ARROW performs comparably to state-of-the-art offline SVR methods, despite not achieving as much variance reduction in the Monte Carlo simulations. This indicates that, even in an offline setting, there is a clear advantage to continuously tracking the alignment statistics, rather than updating them periodically. The gains in accuracy are not always additive (i.e., PSDA + ARROW does not always outperform each method individually), suggesting there could be some more complex interaction mechanisms at play. It was also observed that there is significant sensitivity in the accuracy incurred by the model selection process, which could also explain this phenomenon.

\subsection{Training speed} \label{training speed}

Overall wall-clock training times are reported for each sampler and dataset in Table \ref{times}. These are averaged over all data splits, hyperparameters, and repeats from the previous experiment, including both CORAL and MMD. Again, ARROW performs comparatively to preceding techniques on these mid-scale datasets. Although, being an online technique, we note that the runtime is completely independent of dataset size, so ARROW may be the only viable option on extremely large datasets.

\section{Conclusion}\label{conclusion}

This paper introduced ARROW, the first online SVR method for the MMD and CORAL losses, based on a novel framework of instance reweighting and exponential moving averages. We demonstrate competitiveness with offline methods in terms of variance reduction, speed, and target-domain performance. Future work could include extending this framework to other losses, both for UDA and more generally.

\section{Acknowledgements}\label{acknowledgements}

The author acknowledges the use
of the IRIDIS High Performance Computing Facility, and associated
support services at the University of Southampton, in the completion of
this work.

\bgroup
\setlength{\tabcolsep}{0.5em}
\begin{table*}[t] \scriptsize
\centering
\caption{Average test accuracy for Spawrious by data split.}
\label{spawrious split}
\begin{tabular}{@{}l|cccccc|c@{}}
\toprule
\textbf{Method}    & \textbf{O2O-Easy}   & \textbf{O2O-Medium} & \textbf{O2O-Hard}   & \textbf{M2M-Easy}   & \textbf{M2M-Medium}  & \textbf{M2M-Hard}    & \textbf{Average}    \\ \midrule
ERM                & 68.6 ± 1.7          & 62.6 ± 0.8          & 62.1 ± 0.7          & 70.2 ± 1.8          & 45.0 ± 1.3           & 43.0 ± 1.2           & 58.6 ± 0.5          \\
DANN               & 91.4 ± 3.0          & 57.1 ± 3.5          & 71.1 ± 3.2          & 91.1 ± 0.1          & 54.8 ± 4.4           & 39.8 ± 3.4           & 67.5 ± 1.3          \\
CDAN               & 91.9 ± 1.7          & 57.0 ± 3.0          & 70.3 ± 2.2          & 92.9 ± 0.9          & 58.3 ± 3.8           & 44.3 ± 7.7           & 69.1 ± 1.6          \\
CDAN + SDAT        & 92.9 ± 1.4          & 54.3 ± 3.5          & 73.7 ± 6.6          & 83.3 ± 3.0          & 60.3 ± 4.3           & 53.0 ± 6.0           & 69.6 ± 1.8          \\
CDAN + ELS         & 89.8 ± 1.9          & 58.4 ± 2.1          & 67.3 ± 1.5          & 89.9 ± 2.2          & 62.3 ± 2.3           & 56.1 ± 10.2          & 70.6 ± 1.9          \\
ARM                & 70.2 ± 2.9          & 58.6 ± 2.1          & 60.6 ± 0.2          & 68.7 ± 1.6          & 42.2 ± 1.8           & 41.7 ± 1.0           & 57.0 ± 0.7          \\
MCC                & 87.6 ± 1.9          & 51.0 ± 0.8          & 54.0 ± 6.3          & 77.5 ± 2.4          & 46.4 ± 0.5           & 42.7 ± 1.2           & 59.9 ± 1.2          \\ \midrule
CORAL              & 70.7 ± 2.3          & 58.4 ± 1.9          & 64.1 ± 0.6          & 78.6 ± 1.5          & 54.1 ± 1.2           & 49.2 ± 0.7           & 62.5 ± 0.6          \\
+ k-means++        & 82.8 ± 3.5          & 58.2 ± 2.4          & 61.4 ± 4.1          & 75.5 ± 2.7          & 54.7 ± 2.7           & 48.6 ± 1.1           & 63.5 ± 1.2          \\
+ DPP              & 79.8 ± 3.4          & 59.8 ± 2.3          & 67.8 ± 2.0          & 79.6 ± 2.3          & 58.5 ± 1.4           & 49.4 ± 1.9           & 65.8 ± 0.9          \\
+ Anticlustering   & 89.1 ± 4.7          & 57.6 ± 3.3          & 73.7 ± 6.1          & 87.5 ± 2.2          & 51.4 ± 3.2           & 47.5 ± 2.5           & 67.8 ± 1.6          \\
+ VaRDASS          & 90.1 ± 2.2          & \textbf{62.2 ± 1.0} & \textbf{79.6 ± 2.8} & 77.0 ± 2.7          & 51.7 ± 1.6           & 45.8 ± 0.4           & 67.7 ± 0.8          \\
+ ORDERED          & 88.2 ± 2.2          & 61.6 ± 1.6          & 78.1 ± 3.5          & 84.1 ± 5.0          & 60.5 ± 2.6           & 50.5 ± 2.1           & 70.5 ± 1.3          \\
+ PSDA             & \textbf{92.4 ± 2.2} & 53.7 ± 1.7          & 73.9 ± 7.2          & \textbf{86.0 ± 1.8} & 60.9 ± 1.7           & 54.0 ± 1.3           & 70.2 ± 1.4          \\
+ ARROW            & 85.1 ± 1.4          & 54.7 ± 1.9          & 78.1 ± 4.4          & 79.2 ± 2.5          & 60.8 ± 4.5           & 50.1 ± 0.9           & 68.0 ± 1.2          \\
+ PSDA + ARROW     & 84.7 ± 4.8          & 58.8 ± 2.6          & 79.5 ± 6.1          & 81.9 ± 0.9          & \textbf{65.7 ± 3.9}  & \textbf{64.8 ± 6.8}  & \textbf{72.6 ± 1.9} \\ \midrule
MMD                & 79.2 ± 3.3          & 61.9 ± 1.2          & 65.5 ± 3.4          & 76.2 ± 3.4          & 55.3 ± 3.4           & 48.1 ± 0.7           & 64.4 ± 1.1          \\
+ k-means++        & 83.7 ± 6.3          & 58.6 ± 2.4          & 68.4 ± 4.5          & 79.3 ± 2.7          & 60.0 ± 3.0           & 52.5 ± 4.5           & 67.1 ± 1.7          \\
+ DPP              & 83.6 ± 4.5          & \textbf{62.9 ± 0.9} & 63.5 ± 3.2          & 79.1 ± 3.1          & 57.4 ± 4.0           & 45.6 ± 1.7           & 65.4 ± 1.3          \\
+ Anticlustering   & 74.5 ± 6.0          & 61.7 ± 1.0          & 85.9 ± 5.7          & 75.4 ± 4.1          & 63.7 ± 10.8          & 44.6 ± 3.0           & 67.6 ± 2.4          \\
+ VaRDASS          & \textbf{94.2 ± 1.8} & 61.5 ± 1.4          & 72.7 ± 4.5          & 76.9 ± 4.3          & \textbf{75.9 ± 10.6} & 48.1 ± 5.1           & 71.6 ± 2.2          \\
+ ORDERED          & 93.5 ± 1.3          & 56.4 ± 2.4          & \textbf{85.1 ± 1.9} & \textbf{88.6 ± 0.8} & 70.5 ± 7.8           & 62.1 ± 10.9          & \textbf{76.1 ± 2.3} \\
+ PSDA             & 87.2 ± 6.7          & 62.4 ± 5.5          & 77.4 ± 8.9          & 86.6 ± 5.5          & 60.9 ± 11.2          & \textbf{67.5 ± 13.0} & 73.7 ± 3.6          \\
+ ARROW            & 82.6 ± 7.0          & 56.5 ± 2.2          & 65.8 ± 1.9          & 78.0 ± 6.4          & 61.3 ± 4.1           & 46.8 ± 3.4           & 65.2 ± 1.9          \\
+ PSDA + ARROW     & 84.2 ± 8.3          & 60.0 ± 3.1          & 64.0 ± 3.4          & 82.5 ± 5.2          & 62.9 ± 6.0           & 51.3 ± 4.2           & 67.5 ± 2.2          \\ \bottomrule
\end{tabular}
\end{table*}
\egroup

\begin{table*}[t] \scriptsize
\caption{Average test accuracy for Office-Home by data split.}
\label{office split}
\centering
\begin{tabular}{@{}l|cccccccccccc|c@{}}
\toprule
\textbf{Method}    & \textbf{C-A}  & \textbf{A-C}  & \textbf{P-A}  & \textbf{A-P}  & \textbf{R-A}  & \textbf{A-R}  & \textbf{P-C}  & \textbf{C-P}  & \textbf{R-C}  & \textbf{C-R}  & \textbf{R-P}  & \textbf{P-R}  & \textbf{Average}    \\ \midrule
ERM                & 32.9          & 33.9          & 30.7          & 45.7          & 49.1          & 57.5          & 34.2          & 46.2          & 36.5          & 51.8          & 63.2          & 59.6          & 45.1 ± 0.4          \\
DANN               & 31.1          & 34.9          & 28.9          & 39.9          & 46.9          & 50.5          & 31.7          & 46.8          & 39.7          & 48.7          & 63.4          & 52.7          & 42.9 ± 0.4          \\
CDAN               & 36.8          & 31.2          & 30.0          & 40.5          & 47.3          & 52.9          & 33.3          & 44.4          & 42.6          & 49.3          & 60.7          & 54.8          & 43.7 ± 0.5          \\
CDAN + SDAT        & 36.0          & 36.1          & 30.9          & 41.9          & 48.0          & 54.5          & 37.8          & 45.7          & 43.5          & 50.5          & 66.9          & 55.7          & 45.6 ± 0.2          \\
CDAN + ELS         & 35.1          & 30.7          & 29.3          & 38.3          & 45.7          & 52.9          & 33.6          & 44.3          & 41.6          & 46.7          & 62.7          & 55.2          & 43.0 ± 0.3          \\
ARM                & 34.2          & 31.3          & 30.0          & 43.1          & 49.2          & 56.5          & 32.6          & 46.4          & 35.6          & 47.6          & 63.1          & 56.0          & 43.8 ± 0.2          \\
MCC                & 33.8          & 38.3          & 34.3          & 50.7          & 49.9          & 58.7          & 36.4          & 54.1          & 44.2          & 55.6          & 69.1          & 59.8          & 48.7 ± 0.3          \\ \midrule
CORAL              & 39.4          & 35.2          & 34.7          & 40.8          & 57.1          & 56.8          & 35.8          & 48.3          & 43.2          & 49.5          & 68.9          & 60.0          & 47.5 ± 0.1          \\
+ k-means++        & 41.5          & \textbf{40.3} & 40.0          & 46.1          & 53.3          & 56.7          & 40.1          & 51.0          & 45.6          & 55.7          & 70.6          & 64.1          & 50.4 ± 0.5          \\
+ DPP              & 40.8          & 36.0          & 36.9          & 40.5          & 54.5          & 56.3          & 35.8          & 48.7          & 43.5          & 49.6          & 68.9          & 61.0          & 47.7 ± 0.2          \\
+ Anticlustering   & 39.2          & 37.2          & 36.8          & 40.4          & 56.5          & 56.8          & 38.4          & 49.9          & 45.4          & 54.2          & 70.1          & 62.6          & 49.0 ± 0.2          \\
+ VaRDASS          & 39.7          & 35.6          & 36.6          & 42.8          & 55.7          & 57.9          & 39.3          & 50.3          & 47.3          & 52.8          & 71.6          & 62.3          & 49.3 ± 0.4          \\
+ ORDERED          & 41.9          & 37.3          & 39.6          & 44.9          & 58.6          & 58.3          & 42.2          & 51.4          & 47.4          & 55.8          & 71.5          & 64.0          & 51.1 ± 0.4          \\
+ PSDA             & \textbf{44.2} & \textbf{40.3} & \textbf{41.2} & 46.2          & \textbf{58.7} & 60.0          & \textbf{44.5} & \textbf{54.0} & 50.6          & \textbf{55.9}          & \textbf{71.9} & \textbf{65.1}          & \textbf{52.7 ± 0.4} \\
+ ARROW            & 40.5          & 37.1          & 35.4          & 44.1          & 57.8          & 57.5          & 39.6          & 50.5          & 45.4          & 54.1          & 68.9          & 60.6          & 49.3 ± 0.3          \\
+ PSDA + ARROW     & \textbf{44.2} & 39.7          & 38.9          & \textbf{48.3} & 57.7          & \textbf{60.4} & 42.9          & 53.2          & \textbf{50.7}          & \textbf{55.9}          & 69.1          & 64.0          & 52.1 ± 0.3          \\ \midrule
MMD                & 32.4          & 35.4          & 31.5          & 47.0          & 49.5          & 55.7          & 32.2          & 48.4          & \textbf{41.7} & 49.3          & 66.6          & 56.0          & 45.5 ± 0.3          \\
+ k-means++        & 33.6          & 33.7          & 32.4          & 43.9          & 51.8          & 53.4          & 31.9          & 48.0          & 39.3          & 52.5          & 66.2          & 55.5          & 45.2 ± 0.2          \\
+ DPP              & 31.6          & 34.9          & 31.0          & 45.2          & 51.0          & 56.3          & 33.4          & \textbf{51.2} & 39.2          & 50.1          & \textbf{67.2} & 59.0          & 45.9 ± 0.4          \\
+ Anticlustering   & 35.1          & 33.2          & 33.0          & 44.6          & 50.1          & 57.2          & 33.3          & 48.4          & 36.6          & 51.0          & 63.8          & 58.3          & 45.4 ± 0.5          \\
+ VaRDASS          & 33.8          & 33.5          & 32.4          & 45.8          & 50.4          & 57.0          & 32.5          & 49.8          & 37.7          & 51.8          & 65.3          & 58.8          & 45.7 ± 0.2          \\
+ ORDERED          & 35.9          & 36.4          & 33.2          & \textbf{47.1} & 48.7          & 55.7          & 32.7          & 49.8          & 39.0          & 52.3          & 66.2          & 59.8          & 46.4 ± 0.2          \\
+ PSDA             & 33.5          & 35.1          & 31.6          & 45.8          & 51.8          & \textbf{57.3} & 33.4          & 49.1          & 38.5          & 52.0          & 65.4          & \textbf{61.5} & 46.3 ± 0.3          \\
+ ARROW            & \textbf{36.8} & \textbf{35.8} & 30.3          & 45.2          & 51.4          & 56.0          & 32.6          & 50.5          & 43.9          & \textbf{53.2} & 66.5          & 59.6          & 46.8 ± 0.2          \\
+ PSDA + ARROW     & 36.3          & \textbf{35.8} & \textbf{33.6} & 46.2          & \textbf{52.8} & 57.1          & \textbf{33.8}          & 50.8          & 41.1          & 52.1          & 66.7          & 60.2          & \textbf{47.2 ± 0.3} \\ \bottomrule
\end{tabular}
\end{table*}

\begin{table}[t] \scriptsize
\caption{Average test accuracy for Humpbacks by data split.}
\centering
\label{humps split}
\begin{tabular}{@{}l|cccc|c@{}}
\toprule
\textbf{Method}    & \textbf{Domain 1}   & \textbf{Domain 2}   & \textbf{Domain 3}   & \textbf{Domain 4}   & \textbf{Average}    \\ \midrule
ERM                & 70.3 ± 2.7          & 92.0 ± 1.9          & 78.1 ± 3.0          & 96.2 ± 0.6          & 84.2 ± 1.1          \\
DANN               & 60.5 ± 4.2          & 90.1 ± 2.1          & 63.9 ± 6.1          & 76.1 ± 11.1         & 72.6 ± 3.4          \\
CDAN               & 61.6 ± 4.2          & 82.2 ± 4.7          & 73.5 ± 3.0          & 84.4 ± 0.6          & 75.4 ± 1.7          \\
CDAN + SDAT        & 63.7 ± 3.0          & 81.2 ± 6.7          & 63.6 ± 3.7          & 78.9 ± 3.2          & 71.8 ± 2.2          \\
CDAN + ELS         & 62.7 ± 2.4          & 85.6 ± 3.4          & 70.8 ± 2.2          & 83.9 ± 2.3          & 75.8 ± 1.3          \\
ARM                & 78.8 ± 3.4          & 95.9 ± 1.0          & 72.4 ± 3.6          & 89.6 ± 2.0          & 84.2 ± 1.4          \\
MCC                & 70.0 ± 5.9          & 87.7 ± 4.9          & 76.1 ± 4.1          & 97.1 ± 0.5          & 82.7 ± 2.2          \\ \midrule
CORAL              & 77.2 ± 1.4          & 87.7 ± 3.4          & 83.6 ± 4.2          & 92.7 ± 1.9          & 85.3 ± 1.5          \\
+ k-means++        & 79.1 ± 1.7          & 97.8 ± 0.6          & 85.4 ± 4.6          & 93.8 ± 1.0          & 89.1 ± 1.3          \\
+ DPP              & 81.0 ± 1.9          & 97.4 ± 1.0          & 84.4 ± 5.1          & 94.9 ± 1.2          & 89.5 ± 1.4          \\
+ Anticlustering   & 78.8 ± 1.6          & 95.6 ± 2.0          & 87.0 ± 5.9          & 93.1 ± 2.5          & 88.6 ± 1.7          \\
+ VaRDASS          & 78.4 ± 2.8          & 95.1 ± 1.6          & 85.3 ± 4.2          & 94.4 ± 1.2          & 88.3 ± 1.3          \\
+ ORDERED          & 79.8 ± 1.9          & \textbf{98.0 ± 0.9} & 92.0 ± 3.1          & 93.8 ± 1.8          & 90.9 ± 1.0          \\
+ PSDA             & 80.7 ± 1.4          & 97.5 ± 0.9          & 89.9 ± 2.9          & \textbf{96.0 ± 0.7} & 91.0 ± 0.9          \\
+ ARROW            & \textbf{83.3 ± 0.5} & 97.9 ± 0.3          & 97.4 ± 1.7          & 93.4 ± 1.0          & \textbf{93.0 ± 0.5} \\
+ PSDA + ARROW     & 81.3 ± 0.4          & 97.6 ± 0.4          & \textbf{97.6 ± 1.6} & 93.0 ± 1.1          & 92.4 ± 0.5          \\ \midrule
MMD                & 78.3 ± 2.5          & 95.0 ± 1.2          & 83.3 ± 4.2          & 94.3 ± 1.1          & 87.7 ± 1.3          \\
+ k-means++        & 79.7 ± 1.9          & 97.6 ± 0.4          & 79.5 ± 5.7          & \textbf{96.7 ± 0.3} & 88.4 ± 1.5          \\
+ DPP              & 83.4 ± 1.0          & 97.9 ± 0.6          & 83.4 ± 5.9          & 96.6 ± 0.6          & 90.3 ± 1.5          \\
+ Anticlustering   & 80.5 ± 2.4          & 95.2 ± 0.9          & 92.3 ± 4.7          & 93.8 ± 0.5          & 90.5 ± 1.3          \\
+ VaRDASS          & 79.2 ± 1.1          & \textbf{98.4 ± 0.5} & 95.5 ± 1.2          & 95.0 ± 1.0          & 92.0 ± 0.5          \\
+ ORDERED          & 78.9 ± 2.0          & 98.1 ± 0.3          & 94.8 ± 2.6          & 95.5 ± 1.0          & 91.8 ± 0.9          \\
+ PSDA             & 80.2 ± 2.3          & 97.3 ± 0.7          & 95.3 ± 0.9          & 95.6 ± 0.6          & 92.1 ± 0.7          \\
+ ARROW            & 81.3 ± 0.9          & 97.2 ± 0.5          & \textbf{98.9 ± 0.7} & 93.3 ± 0.7          & \textbf{92.7 ± 0.4} \\
+ PSDA + ARROW     & \textbf{81.5 ± 0.9} & 97.2 ± 0.4          & 98.8 ± 0.8          & 91.7 ± 0.9          & 92.3 ± 0.4          \\ \bottomrule
\end{tabular}
\end{table}

\begin{table}[t]
\caption{Average wall-clock training times by dataset (seconds).}
\label{times}
\centering
\begin{tabular}{@{}l|ccc@{}}
\toprule
\multicolumn{1}{c|}{\textbf{Sampler}} & \textbf{Spawrious} & \textbf{Office-Home} & \textbf{Humpbacks} \\ \midrule
Uniform random                        & 393 ± 9            & 501 ± 18             & 30 ± 1             \\
k-means++                             & 2312 ± 151         & 997 ± 28             & 345 ± 15           \\
DPP                                   & 4123 ± 109         & 1095 ± 32            & 731 ± 19           \\
Anticlustering                        & 1403 ± 24          & 818 ± 25             & 124 ± 2            \\
VaRDASS                               & 2958 ± 21          & 1166 ± 45            & 528 ± 9            \\
ORDERED                               & 3900 ± 83          & 2579 ± 97            & 1944 ± 43          \\
PSDA                                  & 1329 ± 13          & 604 ± 20             & 286 ± 3            \\
ARROW                                 & 756 ± 20           & 1093 ± 31            & 221 ± 6             \\
PSDA + ARROW                          & 1596 ± 25          & 1262 ± 44            & 224 ± 8             \\ \bottomrule
\end{tabular}
\end{table}

{
    \small
    \bibliographystyle{tmlr}
    \bibliography{references}
}

\end{document}